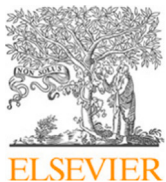
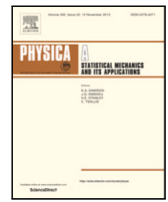

# A Zipf-preserving, long-range correlated surrogate for written language and other symbolic sequences

Marcelo A. Montemurro [a], Mirko Degli Esposti [b],*

[a] *School of Mathematics and Statistics, The Open University, Walton Hall, Milton Keynes, UK*
[b] *Department of Physics and Astronomy, University of Bologna, Bologna, Italy*

A B S T R A C T

Symbolic sequences such as written language and genomic DNA display characteristic frequency distributions and long-range correlations extending over many symbols. In language, this takes the form of Zipf's law for word frequencies together with persistent correlations spanning hundreds or thousands of tokens, while in DNA it is reflected in nucleotide composition and long-memory walks under purine–pyrimidine mappings. Existing surrogate models usually preserve either the frequency distribution or the correlation properties, but not both simultaneously.

We introduce a surrogate model that retains both constraints: it preserves the empirical symbol frequencies of the original sequence and reproduces its long-range correlation structure, quantified by the detrended fluctuation analysis (DFA) exponent. Our method generates surrogates of symbolic sequences by mapping fractional Gaussian noise (FGN) onto the empirical histogram through a frequency-preserving assignment. The resulting surrogates match the original in first-order statistics and long-range scaling while randomising short-range dependencies.

We validate the model on representative texts in English and Latin, and illustrate its broader applicability with genomic DNA, showing that base composition and DFA scaling are reproduced. This approach provides a principled tool for disentangling structural features of symbolic systems and for testing hypotheses on the origin of scaling laws and memory effects across language, DNA, and other symbolic domains.

## 1. Introduction

Human language exhibits complex structure and statistical regularities that emerge from interactions among syntactic, grammatical, and semantic constraints. While traditional linguistic theories emphasise formal rules, quantitative approaches have uncovered robust statistical phenomena in natural texts—such as power-law distributions, scaling laws, and long-range correlations. General statistical regularities in language, such as frequency distributions, scaling relations, and information-theoretic laws, have been widely documented [1–4]. Early work by [5] quantified correlations across letters and sentences using random-walk models. Later, [6] demonstrated fractal long-range correlations in literary corpora via Hurst exponents. Further work by [7] on temporal word-use patterns supports the presence of scaling structures in lexical usage. Recent cross-family analyses covering thousands of languages [8] have confirmed that such scaling laws are pervasive across linguistic typologies.

---

One of the most prominent empirical laws in linguistics is Zipf's law, which states that the frequency $f(r)$ of a word is inversely proportional to its rank $r$ in the frequency-ordered vocabulary [9]:

$$f(r) \sim \frac{1}{r^\gamma}, \quad \gamma \approx 1.$$

This rank-frequency distribution is observed across languages, genres, and modalities, and is often cited as a hallmark of the self-organising properties of language. However, Zipf's law captures only the *first-order statistics* of word usage, ignoring the sequential structure that defines textual meaning and coherence.

Short-range correlations in language largely emerge from syntactic and grammatical constraints operating within the span of sentences. These local dependencies have been modelled and reproduced effectively using Markov processes—for instance, composite generative models that use Hidden Markov Models to account for syntactic word order and dependencies [10]. Indeed, it has been pointed out that these local syntactic correlations underpin sentence-level structure, over which long-range statistical patterns are then layered [6].

Language also exhibits long-range correlations that extend far beyond the scale of the sentence. This temporal organisation in texts has been extensively studied using techniques such as DFA, which was first introduced by Peng et al. for the analysis of DNA sequences [11,12] and later applied to written language. Techniques originally developed for studying correlations in DNA [12,13] were adapted to natural language, revealing that texts exhibit long-range correlations extending over hundreds or thousands of words [5,6,14,15], indicating the presence of statistical dependencies at scales much larger than the sentence. Such correlations are absent in simple generative models such as independent or low-order Markov processes, highlighting the need for models that account for the global structure of language [16].

Surrogate data methods offer a principled framework for probing the origin and significance of such structures. By generating synthetic sequences that preserve selected statistical properties of the original data while randomising others, one can assess which features are responsible for observed phenomena. In the context of language, several surrogate models have been proposed, including random shufflings at the character, word, or sentence levels, as well as Markov and $n$-gram models [5–7,17]. Beyond these approaches, multifractal analyses have been applied to written texts, revealing hierarchical scaling features and complex long-range structures that go beyond simple frequency statistics [16,18]. These surrogates and multifractal techniques help to isolate the role of lexical statistics, syntactic constraints, memory effects, and deeper structural hierarchies in shaping the observed signal.

However, existing surrogate models tend to preserve either the rank-frequency distribution (as in word-level shufflings) or the correlation structure (as in synthetic Gaussian processes), but not both simultaneously. To our knowledge, there is no existing model that generates symbolic surrogate sequences which retain both the empirical Zipf distribution and the long-range correlation structure of the original text.

In this paper, we introduce a new class of surrogate language models that bridges this gap. Starting from an encoded representation of a text, we construct symbolic surrogate sequences that simultaneously:

- Preserve the empirical Zipf distribution of the original text;
- Match its long-range correlation properties, as quantified by the DFA exponent.

Our method maps a long-range correlated real-valued process (FGN) onto a discrete symbolic sequence using a frequency-constrained assignment procedure. The resulting surrogates preserve key statistical and dynamical features of real texts, enabling controlled studies of how lexical statistics and temporal structure shape language.

The paper is structured as follows. In Section 2, we review the empirical evidence for long-range correlations in natural language and the use of DFA to characterise them. Section 3 discusses symbolic encodings for language analysis and motivates the use of Zipf-rank encoding as the foundation for our surrogate model applied to language sequences. Section 4 briefly reviews existing surrogate models and their limitations. Section 5 presents our proposed method, including its mathematical formulation and computational implementation. In Section 6, we illustrate the broader applicability of the surrogate by applying it to genomic DNA. We conclude in Section 7 with a discussion.

## 2. Long-range correlations in language

While the statistical distribution of word frequencies in natural language is well captured by Zipf's law, this law is fundamentally a *static* property: it describes how often different words occur, but says nothing about the *order* in which they appear. Yet, sequential organisation is central to language, it enables coherent syntax, narrative structure, and thematic development. To understand the generative and cognitive underpinnings of language, it is essential to study temporal dependencies and the statistical structure of word order over multiple scales.

At short timescales, linguistic dependencies are governed by local syntactic and lexical constraints. These are well captured by finite-order Markov models, in which the probability of a word depends only on a fixed number of preceding tokens [19–21].

Beyond these short-range effects, language exhibits correlations over much longer timescales, reflecting discourse-level coherence and thematic memory. Early work [5] demonstrated that the distribution of symbol transitions in literary texts displays scaling behaviour that cannot be explained by random or short-memory processes. Schenkel et al. [14] extended these findings using DFA applied to character level encodings of texts, revealing that the probability of observing a given token at time $t$ remains statistically dependent on occurrences at distant positions $t'$, even when $|t - t'|$ thousands of characters (corresponding to hundreds of words in typical prose). Beyond monofractal scaling quantified by a single DFA exponent, written texts can exhibit multifractal structure.

Multifractal detrended fluctuation analysis (MFDFA) has been applied to a variety of linguistic encodings, revealing broad spectra of scaling exponents and suggesting a cascade-like hierarchical organisation in some texts. Recent work by Stanisz et al. [22] presents a comprehensive review of such phenomena and documents both universal and system-specific aspects of multifractality in natural language. In the present work we deliberately restrict attention to monofractal long-range correlations described by a single DFA exponent $\alpha$.

These results support the view that natural language is not only statistically structured at the local level but also exhibits *long-range correlations*, characteristic of complex systems with memory and hierarchical organisation. Identifying and quantifying such correlations is therefore critical for understanding the underlying processes that generate coherent linguistic sequences.

*2.1. Detrended Fluctuation Analysis*

Detrended Fluctuation Analysis is a robust technique for detecting long-range correlations in noisy and potentially nonstationary time series [12,23,24]. Given a discrete time series $x(t)$, the method proceeds by first constructing the cumulative sum, or *profile*, defined as

$$Y(t) = \sum_{i=1}^{t} [x(i) - \bar{x}],$$

where $\bar{x}$ denotes the mean of the series. The profile $Y(t)$ is then divided into non-overlapping segments of equal length $L$. Within each segment, a local polynomial fit (typically linear, but higher-order detrending is possible [23]) is used to estimate the local trend, which is subtracted from the data. The root-mean-square fluctuation from the trend is computed as

$$F(L) = \left\{ \frac{1}{N} \sum_{t=1}^{N} \left[ Y(t) - Y_{\text{fit}}(t) \right]^2 \right\}^{1/2}.$$

This procedure is repeated for various window sizes $L$, and the fluctuation function $F(L)$ is plotted against $L$ on a log–log scale. A power-law relationship,

$$F(L) \sim L^{\alpha},$$

indicates scale-invariant correlations, with the exponent $\alpha$ capturing the nature of the temporal structure.

The DFA exponent $\alpha$ serves as a quantitative indicator of correlation. Values around $\alpha = 0.5$ suggest uncorrelated white noise, while $\alpha < 0.5$ indicates anti-persistent behaviour—where an increase is likely to be followed by a decrease, and vice versa. In contrast, $\alpha > 0.5$ reflects persistent or long-range correlations, as found in many natural and complex systems.

The exponent $\alpha$ is closely related to other descriptors of correlation. For a stationary signal with a power-law decay of autocorrelation, $C(s) \sim s^{-\zeta}$ (with $0 < \zeta < 1$), the relationship is given by

$$\alpha = 1 - \frac{\zeta}{2}.$$

Similarly, for stationary processes with a power spectral density $S(f) \sim f^{-\beta}$, the exponents satisfy

$$\beta = 2\alpha - 1.$$

Thus, white noise corresponds to $\alpha = 0.5$ and $\beta = 0$, while long-range correlations imply $\alpha > 0.5$ and $\beta > 0$ [25].

These scaling relations establish DFA as a bridge between time-domain and frequency-domain characterisations of temporal structure. In particular, they provide a unifying framework for interpreting power-law behaviour in terms of both autocorrelation and spectral properties.

Empirical studies of natural language consistently yield DFA exponents in the range $\alpha \approx 0.6$–$0.8$ when applied to appropriate encodings [5,6,14,15]. These values indicate the presence of long-range memory not captured by Markovian or strictly local models, and appear robust across languages and textual corpora.

## 3. Encoding language for time series analysis

Quantitative analysis of language as a complex system requires transforming symbolic data into numerical sequences suitable for time series methods. The choice of encoding determines which statistical and dynamical features are retained, from surface-level frequency patterns to higher-order syntactic and semantic structures.

Early work focused on character-level representations. In one approach, each character was mapped to its ASCII code, producing an integer-valued time series subsequently analysed with DFA [14]. Related studies used reduced alphabets—e.g., mapping the 32 most frequent characters to integers (0–31) or 5-bit codes—to compress the symbol set [17].

The resulting sequences exhibited power-law scaling in their autocorrelation and spectral properties. An alternative scheme assigned each frequent character a binary time series, indicating its presence or absence at each time step [5]. These sequences generated parallel random walks whose scaling behaviour reflected persistent correlations not attributable to simple frequency effects. While conceptually straightforward, character-level encodings fragment linguistic units and obscure deeper structural dependencies.

To address this limitation, later studies adopted word-level representations. Sequences of word indices or grammatical classes have been used to retain syntactic and lexical organisation [26,27], though these mappings are often heuristic and not easily generalisable.

Alternative schemes based on frequency or word length have also been proposed. In particular, texts have been transformed into frequency time series (FTS), where each word is replaced by its global frequency, or into length time series (LTS), where each word is represented by its number of characters [18,28]. These representations have been analysed using rank-frequency statistics and nonlinear time series techniques such as the Grassberger–Procaccia algorithm [29,30]. Results indicate that correlation detection is highly sensitive to the choice of encoding, emphasising the need for principled mapping strategies.

### 3.1. Zipf-rank encoding

To preserve core statistical properties while avoiding arbitrary mappings, symbolic texts can be transformed into numerical sequences via the *Zipf-rank encoding* [6]. Let $T = \{w(t)\}_{t=1}^{N}$ be a sequence of word tokens drawn from a vocabulary $\mathcal{A} = \{a_1, a_2, \ldots, a_V\}$. The empirical frequency $f(a_i)$ of each word defines a rank ordering:

$$f(a_1) \geq f(a_2) \geq \cdots \geq f(a_V),$$

such that

$$\sum_{i=1}^{V} f(a_i) = N,$$

and each token is assigned its rank $r(t) = \text{rank}(w(t))$, leading to a numerical sequence $R = \{r(t)\}_{t=1}^{N}$.

This encoding preserves the Zipfian distribution of word usage, abstracts away lexical identity, and reduces dimensionality. It also supports a dynamical interpretation: the sequence $R$ defines a discrete-time process on the positive integers, where rank differences $\Delta r(t) = r(t+1) - r(t)$ form a *rank-jump process*. When analysed using DFA, such sequences typically give scaling exponents $\alpha > 0.5$, indicating long-range correlations. In contrast, surrogate sequences obtained by word shuffling collapse to $\alpha \approx 0.5$, reflecting the loss of temporal structure.

Analyses of written language have shown that punctuation marks also display Zipf-like rank–frequency behaviour and that including them as tokens can improve the scaling properties of the resulting symbolic sequences. Recent studies have reported that punctuation contributes systematically to the statistical organisation of texts and can enhance the stability of scaling laws extracted through DFA-based methods [22,31]. In the present work we follow the standard approach in the DFA literature and remove punctuation during preprocessing, thereby isolating lexical contributions to Zipf's law and long-range correlations. However, our surrogate construction is fully compatible with punctuation-aware encodings: punctuation marks may be included as additional symbols in the alphabet, in which case their empirical frequencies and their contribution to long-range correlations are preserved by the surrogate in exactly the same manner as for lexical items.

In this work, the Zipf-rank encoding serves as the basis for surrogate modelling. Its ability to preserve frequency structure while exposing dynamical correlations makes it particularly suited to capturing the dual statistical and temporal nature of natural language.

## 4. Surrogate models of language

Surrogate data methods provide a powerful tool for analysing complex systems by allowing researchers to test hypotheses about the origin of observed statistical properties. The central idea is to construct alternative sequences—*surrogates*—that preserve some features of the original data while randomising others. In the context of language, surrogates are often used to investigate which aspects of linguistic structure are responsible for empirical patterns such as Zipf's law or long-range correlations.

In this section, we briefly review existing surrogate models for language, highlighting their design principles, the properties they preserve, and the limitations that motivate the development of more sophisticated approaches.

### 4.1. Hierarchical shuffling schemes

One of the simplest and most widely used surrogate techniques is *random shuffling* of the original text at different levels of linguistic organisation. These models preserve the identity and frequency of the elements being shuffled, while disrupting their order and hence their temporal structure.

- **Character-level shuffling** destroys all higher-order linguistic information, including word structure and syntax. While it retains character frequencies, it removes Zipf's law and eliminates all correlations beyond the unigram level.
- **Word-level shuffling** preserves the empirical word frequency distribution, and hence Zipf's law, but breaks all syntactic and semantic dependencies. As shown in previous studies [5,6], the DFA exponent of such sequences typically collapses to $\alpha \approx 0.5$, indicating a loss of long-range correlations.
- **Sentence-level shuffling** maintains the internal structure of individual sentences but destroys discourse-level coherence. These surrogates preserve local grammatical patterns and retain Zipf's law, but they substantially reduce the DFA exponent, indicating that sentence ordering contributes significantly to long-range memory [7,15].

Random shuffling separates frequency from correlation by preserving the former while eliminating the latter. Combined with existing surrogates, our method extends this approach, enabling a hierarchical analysis of their independent and joint contributions to language's statistical structure.

### 4.2. Markov and N-gram models

A more structured class of surrogates is based on probabilistic models such as Markov chains and *n*-gram models. In these models, the probability of observing a token depends on a fixed number of previous tokens:

$$P(w_t \mid w_{t-1}, \ldots, w_{t-n+1}).$$

Such models capture short-range dependencies and are widely used in natural language processing (NLP). However, their ability to reproduce global statistical features is limited.

Low-order Markov models can produce approximate Zipf-like frequency distributions under certain assumptions [32], but they fail to reproduce the empirical scaling and do not generate long-range correlations. Higher-order models may approximate Zipfian statistics by overfitting, but they do so without offering insight into the underlying mechanisms. Moreover, these models tend to suffer from state-space explosion and data sparsity, making them impractical for capturing large-scale correlations across long texts.

### 4.3. Correlation-preserving real-valued surrogates

A standard method to generate surrogate time series that preserve linear correlations (as captured by the power spectrum or autocorrelation function) but randomise nonlinear structure is the Fourier Transform (FT) surrogate method [33]. The idea is to replace the Fourier phases of the original signal with random values while retaining the amplitudes, thus generating a linear stochastic process with the same power spectrum as the original.

While FT surrogates preserve second-order statistics, they do not replicate the amplitude distribution of the original data. To overcome this, amplitude-adjusted Fourier transform (AAFT) surrogates were introduced [34], followed by iterative versions (IAAFT) that improve accuracy in matching both the spectrum and amplitude histogram.

For signals displaying long-range correlations ($1/f$-type noise), FGN provides a natural linear model with well-defined second-order statistics governed by the Hurst exponent $H$ [35,36]. Surrogates based on FGN are often used as null models to assess whether the correlations in a real-world time series are consistent with a Gaussian linear process [37]. These processes are stationary and can be efficiently generated using the Fourier filtering method or wavelet-based approaches.

Such correlation-preserving surrogates have been widely applied to climatological, physiological, and geophysical time series [38], where long-memory and power-law correlations coexist with potential nonlinearity and nonstationarity. Their use is essential for distinguishing between linear stochastic dynamics and truly nonlinear mechanisms.

## 5. Zipf-preserving long-range correlated surrogates: mathematical formulation

The models discussed above reveal a consistent trade-off: surrogate sequences typically preserve either the rank–frequency distribution, as in word-level shuffling, or the long-range correlation structure, as in real-valued stochastic processes, but not both. To our knowledge, no existing approach generates symbolic sequences that retain both the empirical Zipf distribution of word frequencies and the long-range correlation properties of the original text, as quantified by DFA.

This gap motivates the development of a new class of surrogates that integrate both constraints in a principled way. By combining the Zipf-rank encoding of symbolic sequences introduced in Section 3.1 with the dynamical structure of long-memory processes, we construct symbolic surrogates that faithfully reproduce the lexical and temporal organisation of natural language.

Our goal is to construct a *surrogate symbolic sequence* $S = \{s(t) \in \mathcal{A} \mid t = 1, \ldots, N\}$ such that:

1. The empirical frequencies $f_S(a_i)$ in $S$ match exactly those of the original sequence $T$:

   $$f_S(a_i) = f(a_i), \quad \forall i = 1, \ldots, V,$$

   thereby preserving Zipf's law;
2. The sequence $S$ exhibits long-range correlations consistent with a prescribed DFA exponent $\alpha$, matching that of the original rank-encoded sequence.

To achieve this, we construct $S$ by *discretising a continuous-valued stochastic process* $Z = \{z(t) \in \mathbb{R} \mid t = 1, \ldots, N\}$, where $Z$ is a realisation of FGN with a specified Hurst exponent $H \in (0.5, 1)$. The FGN process $Z$ is characterised by long-range correlations such that the scaling exponent $\alpha$ obtained via DFA is related to the Hurst exponent by $\alpha = H$.

Let $P(z)$ denote the probability density function of the process $Z$. For FGN, $P(z)$ is Gaussian and stationary. Given that $Z$ consists of continuous, distinct values $\{z(t)\}$, we define a deterministic mapping $\mathcal{M} : \mathbb{R} \to \mathcal{A}$ that assigns each value $z(t)$ to a symbol $a_i \in \mathcal{A}$ such that the resulting discrete process $S = \{s(t) = \mathcal{M}(z(t))\}$ preserves the empirical frequencies $f(a_i)$.

To construct such a mapping, we partition the real line into $V$ disjoint intervals $I_i = (z_{i-1}, z_i]$, with $z_0 = -\infty$, $z_V = +\infty$, and define the endpoints $z_i$ so that

$$\int_{z_{i-1}}^{z_i} P(z)\, dz = \frac{f(a_i)}{N}, \quad \text{for } i = 1, \ldots, V,$$

in such a way that each interval $I_i$ contains a portion $f(a_i)/N$ of the probability mass, as illustrated in Fig. 1.

This procedure ensures that each symbol $a_i$ is assigned to a contiguous quantile range $f(a_i)/N$ of the total values in $Z$, preserving the empirical Zipf distribution. In computational terms, this mapping is implemented by sorting the values $z(t)$, assigning symbols to contiguous blocks according to their empirical frequencies, and then applying the inverse permutation to restore temporal order. This yields a surrogate $S$ that preserves Zipf's law and inherits the correlation structure from $Z$.

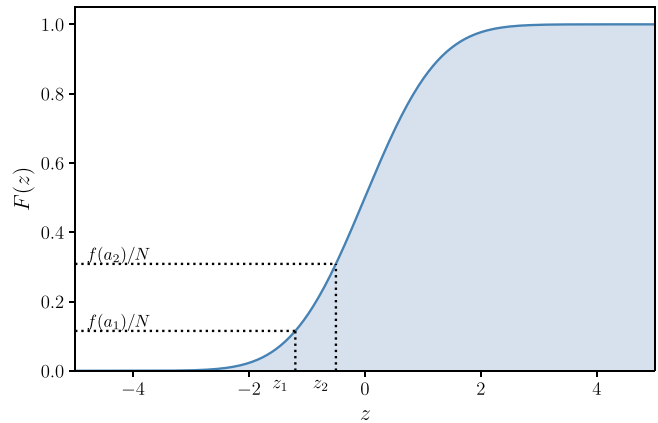

**Fig. 1. Cumulative distribution for the continuous-valued long-range correlated process.** The underlying process is a FGN with a given value of $\alpha$. The $y$-axis is divided according to the normalised Zipf-ordered frequencies. This in turn leads to a partition of the $z$-axis that defines ranges of $z$-values that will be assigned to each of the discrete symbols $a_i$ in order to discretise the process. The figure shows as an example the partitions defined for the first two most frequent words, $a_1$ and $a_2$.

Finally, by applying the inverse permutation that restores the original temporal order, we obtain the surrogate sequence $S$ that inherits the *rank structure* (Zipf's law) from the original text and the *temporal correlation structure* from the continuous process $Z$. However, due to the discretisation inherent in mapping continuous values to symbolic tokens, some correlation is lost, and the DFA exponent $\alpha_S$ of the surrogate sequence is typically lower than that of the underlying FGN.

To match a desired target DFA exponent $\alpha$, we introduce a *correction mechanism* based on a bisection search over the input Hurst exponent $\alpha_0$ of the FGN. By iteratively adjusting $\alpha_0$ and measuring the DFA exponent $\alpha_S$ of the resulting surrogate, we converge to a sequence $S$ satisfying both conditions:

- $f_S(a_i) = f(a_i)$
- $\alpha_S = \alpha$

As a consequence of this construction, the surrogate retains only first-order symbol frequencies and long-range second-order correlations, while higher-order and short-range structures are suppressed. This defines a principled method for generating *Zipf-preserving, long-range correlated symbolic surrogates*. In the next subsection, we describe the practical implementation of this method and the computational algorithms involved.

### 5.1. Computational implementation

We implement the CDF-based mapping via a rank-based procedure that requires only sorting and a deterministic assignment.

*Rank-based discretisation.* Let $Z = \{z(t)\}_{t=1}^N$ be an FGN realisation with input exponent $\alpha_0$, and let $\pi$ be the permutation that sorts $Z$ in ascending order:

$$z(\pi(1)) \leq \cdots \leq z(\pi(N)).$$

Construct $R$ by concatenating the indices $i = 1, \ldots, V$ in order of decreasing frequency, each repeated $f(a_i)$ times. This ensures that the most frequent symbols occupy the top ranks of the sorted FGN. In this way, symbols are assigned by rank, after which the temporal order is restored:

$$s(t) \leftarrow a_{R[\pi^{-1}(t)]}, \qquad t = 1, \ldots, N.$$

This implements the CDF partition without estimating $P(z)$ and guarantees $f_S(a_i) = f(a_i)$ exactly. The measured DFA exponent $\alpha_S$ of $S$ typically underestimates $\alpha_0$ due to discretisation. The computational complexity of the method is $O(N \log N)$, dominated by the sorting step.

*Matching a target exponent.* To achieve a prescribed DFA exponent $\alpha$, we perform a bisection search on $\alpha_0$: for each candidate, generate $S$, measure $\alpha_S$, and update the bracket until $|\alpha_S - \alpha| < \epsilon$. If needed, we reseed the generator to avoid stagnation. The result is a surrogate $S$ that matches Zipf frequencies exactly and the long-range correlation exponent within tolerance.

The following pseudocode summarises the rank-based surrogate generation procedure.

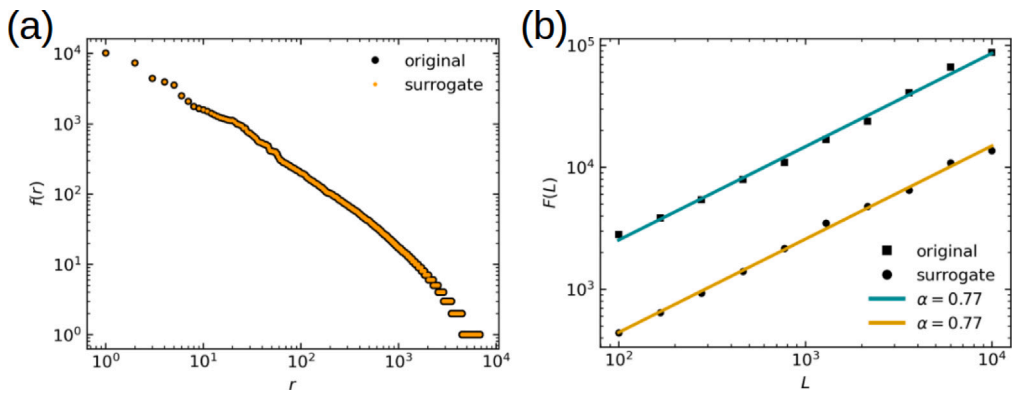

**Fig. 2. Frequency distribution and long-range correlations in surrogate texts**. (a) comparison of the Zipf's distribution for Darwin's On the origin of species (English) and its long-range correlated surrogate. (b) Comparison of long-range correlations between On the origin of species and its long-range correlated surrogate.

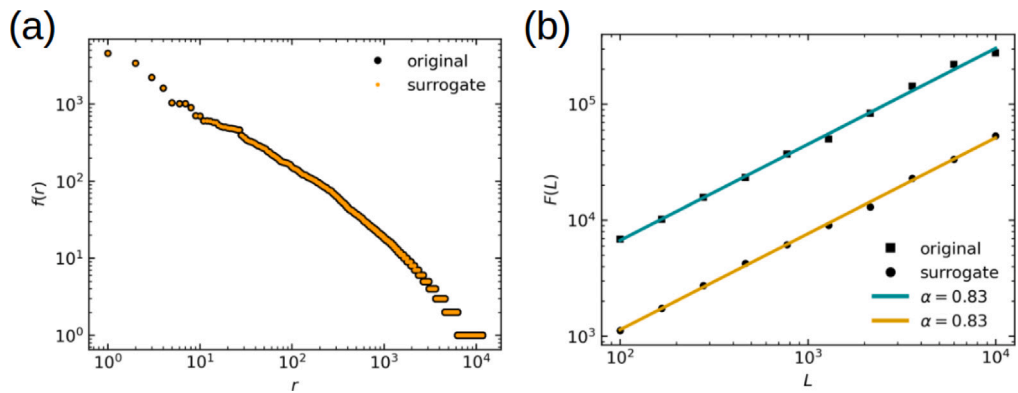

**Fig. 3. Frequency distribution and long-range correlations in surrogate texts**. (a) comparison of the Zipf's distribution for Newton's Principia Mathematica (Latin) and its long-range correlated surrogate. (b) Comparison of long-range correlations between Newton's Principia Mathematica (Latin) and its long-range correlated surrogate.

---

**Algorithm 1** Generate Zipf-preserving long-range correlated surrogate
---
**Require:** $\alpha_0$ (FGN input exponent), $T = \{w(t)\}$ (original text), $s$ (seed)
**Ensure:** $S = \{s(t)\}$ (surrogate sequence)
1: $N \leftarrow |T|$; extract vocabulary $\mathcal{A} = \{a_1, \ldots, a_V\}$ and frequencies $f(a_i)$
2: Initialise RNG with seed $s$ if reproducibility is required
3: Generate FGN $Z = \{z(t)\}_{t=1}^{N}$ with exponent $\alpha_0$
4: Compute sorting permutation $\pi$ s.t. $z(\pi(1)) \leq \cdots \leq z(\pi(N))$; compute inverse $\pi^{-1}$
5: Build list $R$ by concatenating each index $i$ (corresponding to $a_i$) exactly $f(a_i)$ times, in order of decreasing frequency
6: **for** $t = 1$ to $N$ **do**      $s(t) \leftarrow a_{R[\pi^{-1}(t)]}$      **end for**
7: **return** $S$

---

This computational approach generates surrogate symbolic sequences suitable for comparative studies of scaling behaviour, validation of correlation analyses, and the testing of hypotheses concerning the structure of natural language and other symbolic processes.

Two examples of the application of the surrogate to real texts can be seen in Figs. 2 and 3. In both cases, panel (a) shows the close agreement between the Zipf distributions of the original and surrogate texts, while panel (b) compares their DFA analyses, revealing identical fluctuation exponents within the scale range used to generate the surrogate.

## 6. Application to other symbolic sequences: DNA

Genomic sequences also exhibit long-range correlations and non-trivial compositional structure, making them a natural testbed for assessing the broader applicability of our surrogate model. To demonstrate the generality of the framework, we applied it to genomic DNA—a canonical example of symbolic data with well-established long-range correlations. In language, Zipf-rank encoding provides an efficient mapping that preserves the strongly skewed frequency distribution while producing a numerical series suitable for DFA analysis. However, this encoding is not essential to the surrogate method itself: for symbolic systems with small alphabets, such as DNA, simpler direct encodings suffice.

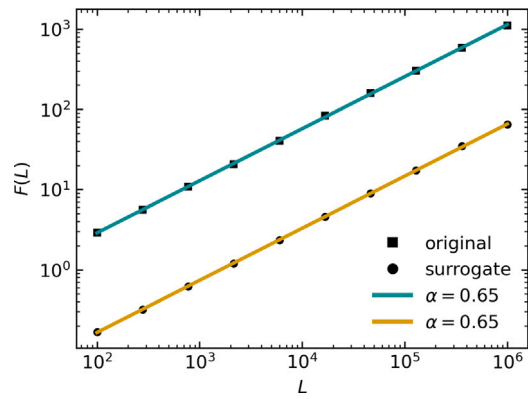

**Fig. 4. Long-range correlations in DNA.** DFA of *Drosophila melanogaster* chromosome 2L (squares) and its long-range surrogate (circles) under the purine–pyrimidine mapping ($\{A, G\} \to +1$, $\{C, T\} \to -1$). Both sequences exhibit the same scaling $F(L) \propto L^\alpha$ with $\alpha = 0.65$ over $10^2 \leq L \leq 10^6$. By construction, the surrogate preserves the exact base composition of the original (A: 29.1%, C: 20.9%, G: 20.9%, T: 29.1%).

Following Peng et al. [11,12], we used the purine–pyrimidine (R/Y) mapping, assigning

$$\{A, G\} \to +1, \qquad \{C, T\} \to -1.$$

This transformation produces a numerical series suitable for DFA, and has been shown to reveal persistent correlations extending over scales of $10^2$ to $10^6$ bases in a wide range of genomes [13,32].

We analysed chromosome 2L of *Drosophila melanogaster* (NCBI accession AE014134.6). The DFA of the purine–pyrimidine walk yielded a scaling exponent $\alpha \approx 0.65$ over scales from $10^2$ to $10^6$ bases, consistent with previous reports of long-range persistence in genomic DNA. Using the proposed surrogate method with an alphabet restricted to $\{A, C, G, T\}$ and the empirical nucleotide frequencies of the chromosome, we generated a Zipf-preserving long-range correlated surrogate sequence. By construction, the surrogate matched the base composition of the original sequence exactly. Its DFA profile was virtually indistinguishable from that of the natural sequence, yielding the same scaling exponent of $\alpha \approx 0.65$.

As expected, higher-order structures such as dinucleotide frequencies were not preserved, reflecting the fact that the surrogate retains only first-order statistics and long-range second-order correlations. This mirrors our findings in natural language: the surrogate provides a principled null model that preserves frequency structure and long-range memory while randomising local dependencies. Fig. 4 shows the DFA scaling curves for the original and surrogate DNA sequences, illustrating their overlap. In a broader genomic context, such surrogates may be used as realistic null models when testing for non-random organisation of genes or regulatory elements, since they preserve both base composition and long-range memory while removing local sequence constraints.

## 7. Discussion and conclusions

In this work, we have introduced a new class of symbolic surrogates that simultaneously preserve first-order frequency statistics and long-range correlation properties of written language. Building on Zipf-rank encoding of texts and long-memory processes such as FGN, our method generates sequences that reproduce both the empirical Zipf distribution and the scaling exponent obtained with DFA. By construction, these surrogates randomise short-range dependencies such as syntactic and lexical patterns, while maintaining the dual constraints of frequency distribution and long-range temporal organisation. In this sense the model is explicitly a linear, stationary null model for symbolic sequences: it isolates the contribution of long-memory second-order statistics against a background where higher-order structure has been deliberately removed. This fills a methodological gap left by previous approaches, which typically preserve either frequency structure (e.g., word-level shuffling [5,6]) or correlation properties (e.g., Fourier-transform or FGN surrogates [33,34,37]), but not both simultaneously. Conceptually, our approach bears a loose analogy to amplitude-adjusted Fourier transform (AAFT) surrogates [34], in that both reconcile a target correlation structure with a prescribed marginal distribution through iterative remapping. However, whereas AAFT operates on continuous-valued signals and preserves amplitude histograms, our method applies to symbolic sequences and enforces empirical frequency distributions through mapping of long-memory Gaussian processes onto discrete symbol sets.

The results obtained for representative texts in English and Latin show that the surrogates reproduce the Zipf distribution exactly and match the DFA scaling exponent of the original sequences. This demonstrates that a substantial part of the long-range structure of language can be captured at the level of second-order statistics once lexical frequencies are controlled for. At the same time, differences between the original and surrogate texts at the level of higher-order organisation—including syntax, semantics, and discourse—remain evident, indicating that these structures contribute to long-range organisation beyond second-order correlations. The surrogate framework therefore provides a principled null model for decomposing the contributions of lexical statistics, long-range correlations, and higher-order linguistic features.

To illustrate the wider applicability of the method, we applied it to genomic DNA. Using a purine–pyrimidine mapping [11–13], we showed that the surrogate preserves both the base composition and the DFA scaling exponent of chromosome 2L of *Drosophila melanogaster*. As in the linguistic case, short-range sequence features such as dinucleotide statistics are not preserved, highlighting the specificity of the surrogate to first-order distributions and long-range correlations. This example shows that the approach is not restricted to language but extends naturally to other symbolic domains where frequency biases and long-range dependencies coexist.

Our surrogates are constructed from fractional Gaussian noise and are therefore monofractal by design: they reproduce a prescribed DFA exponent $\alpha$ but cannot generate non-trivial multifractal spectra. When empirical texts or genomes exhibit multifractality under MFDFA [22], discrepancies between their multifractal spectra and those of the surrogate ensemble provide a principled way to attribute such higher-order scaling to genuinely nonlinear, hierarchical, or nonstationary mechanisms beyond the linear long-memory component captured here.

More broadly, the surrogate framework offers a powerful tool for hypothesis testing in systems where scaling laws and memory effects are thought to emerge from the interplay between local constraints and global organisation [6,7,39]. By providing controlled null models, it becomes possible to disentangle the statistical contributions of frequency distributions, long-range correlations, and higher-order structures to the observed signal. This can help clarify, for instance, whether fluctuations in vocabulary growth [40] or genomic variability arise primarily from second-order correlations or from higher-order structural organisation.

The surrogate framework developed here is primarily methodological, in the tradition of surrogate-data approaches used to separate linear long-range correlations from higher-order structure. Nevertheless, it enables a range of practical applications in symbolic systems. In quantitative linguistics, ensembles of Zipf-preserving, long-range correlated surrogates can serve as null models for assessing how much of the scaling behaviour of a text arises from lexical statistics and linear long-memory alone, and how much must be attributed to higher-order linguistic organisation. Similar ideas apply to the analysis of genomic sequences, where surrogates that preserve base composition and long-range correlations but randomise local structure provide realistic null hypotheses for detecting non-random motif arrangements or regulatory patterns. Comparable uses may be explored in musical sequences, financial records, and other symbolic representations exhibiting scaling behaviour. We expect such domain-specific applications to be developed in future work.

In conclusion, we have proposed and validated a surrogate modelling framework that simultaneously preserves empirical frequency distributions and long-range correlations in symbolic sequences. Our results for natural language and genomic DNA show that the method captures fundamental scaling features shared by diverse symbolic systems, while preserving the potential to detect additional structure beyond second-order statistics. This dual capability makes the surrogate a valuable methodological tool for future investigations of scaling laws and memory effects in language, DNA, and other complex symbolic systems. Future work may extend the framework to other domains—such as musical notation, financial time series, or code repositories—to examine how domain-specific constraints interact with universal statistical regularities.

**CRediT authorship contribution statement**

**Marcelo A. Montemurro:** Writing – review & editing, Conceptualization. **Mirko Degli Esposti:** Writing – review & editing.

**Declaration of competing interest**

The authors declare that they have no known competing financial interests or personal relation-ships that could have appeared to influence the work reported in this paper.

**Acknowledgements**

M.A.M. acknowledges the support of the School of Mathematics and Statistics, The Open University, UK. M.D.E. acknowledges the support of the Department of Physics, University of Bologna, Italy.

**Data availability**

Data will be made available on request.